\documentclass[letterpaper, 10 pt, conference]{ieeeconf} 

\IEEEoverridecommandlockouts

\usepackage[utf8]{inputenc}
\usepackage{amsfonts}
\usepackage{float}
\usepackage{stfloats}  
\usepackage[T1]{fontenc}
\usepackage[noadjust]{cite}

\usepackage [autostyle, english = american]{csquotes}
\MakeOuterQuote{"}
\usepackage{textcomp}
\usepackage{amssymb}
\usepackage{placeins}
\usepackage{blindtext}
\usepackage{graphicx}
\usepackage{lettrine}
\usepackage{xcolor}
\usepackage{lipsum}
\usepackage[fleqn]{amsmath}
\usepackage{multicol}
\usepackage{multirow}
\usepackage{textcomp}
\usepackage{layouts}
\usepackage[font=small,skip=3pt]{caption}
\usepackage{amsmath}
\usepackage{color}
\usepackage{mathtools}
\usepackage{booktabs}

\usepackage{array}
\newcolumntype{L}[1]{>{\raggedright\let\newline\\\arraybackslash\hspace{0pt}}m{#1}}
\newcolumntype{C}[1]{>{\centering\let\newline\\\arraybackslash\hspace{0pt}}m{#1}}
\newcolumntype{R}[1]{>{\raggedleft\let\newline\\\arraybackslash\hspace{0pt}}m{#1}}
\usepackage[linesnumbered]{algorithm2e}
\usepackage{bbm}
\usepackage{duckuments}
\usepackage[colorlinks=true, linkcolor=blue, citecolor=blue, urlcolor=blue]{hyperref}
\usepackage[noadjust]{cite}
\usepackage{subcaption}
\usepackage{array}
\newcolumntype{P}[1]{>{\centering\arraybackslash}p{#1}}
\usepackage{balance}
\title{\LARGE \bf

Fixed-Time Dynamic Landing of Quadrotors using Adaptive Unscented Kalman Filtering and Nonlinear Model Predictive Control

}
\author{Mohammadreza Izadi, Zeinab Shayan, Steven Waslander, Reza Faieghi
\thanks{M. Izadi, Z. Shayan, and R. Faieghi are with the Autonomous Vehicles Laboratory, Department of Aerospace Engineering, Toronto Metropolitan University, Toronto, Canada. {\tt\small \{mizadi, zshayan, reza.faieghi\}@torontomu.ca}. 
S. Waslander is with the University of Toronto Institute for Aerospace Studies, University of Toronto, Toronto, Canada. {\tt\small steven.waslander@utoronto.ca}.}
}

\begin{document}
\maketitle
\thispagestyle{empty}
\pagestyle{empty}
\bstctlcite{IEEEexample:BSTcontrol}

\begin{abstract}
This paper introduces an estimation and control framework for dynamic landing of multi-rotor uncrewed aerial vehicles on moving platforms. The proposed method integrates nonlinear model predictive control with a real-time minimum-jerk trajectory planner that enforces a prescribed touchdown time, enabling consistent timing during the terminal descent. To enhance robustness in the presence of time-varying sensing quality, we utilize an adaptive unscented kalman filter that updates the process and measurement noise statistics online. In addition, we provide a reference feasibility analysis showing that minimum-jerk references induce bounded thrust and torque commands under standard tracking hypotheses. The proposed framework is evaluated in simulation and hardware experiments, and it is shown to achieve repeatable landings and improved platform velocity prediction accuracy relative to EKF/UKF-based methods.
\end{abstract}


\section{INTRODUCTION}\label{se:intro}
\subsection{Motivation}
The ability to land autonomously on moving platforms such as ground vehicles and maritime vessels increases the operational efficiency and expands the deployment scenarios of multi-rotor uncrewed aerial vehicles (MRUAVs). 

\begin{figure}[t]
    \centering
    \includegraphics[trim={4cm, 0cm, 4cm, 0.9cm}, clip, width = \linewidth]{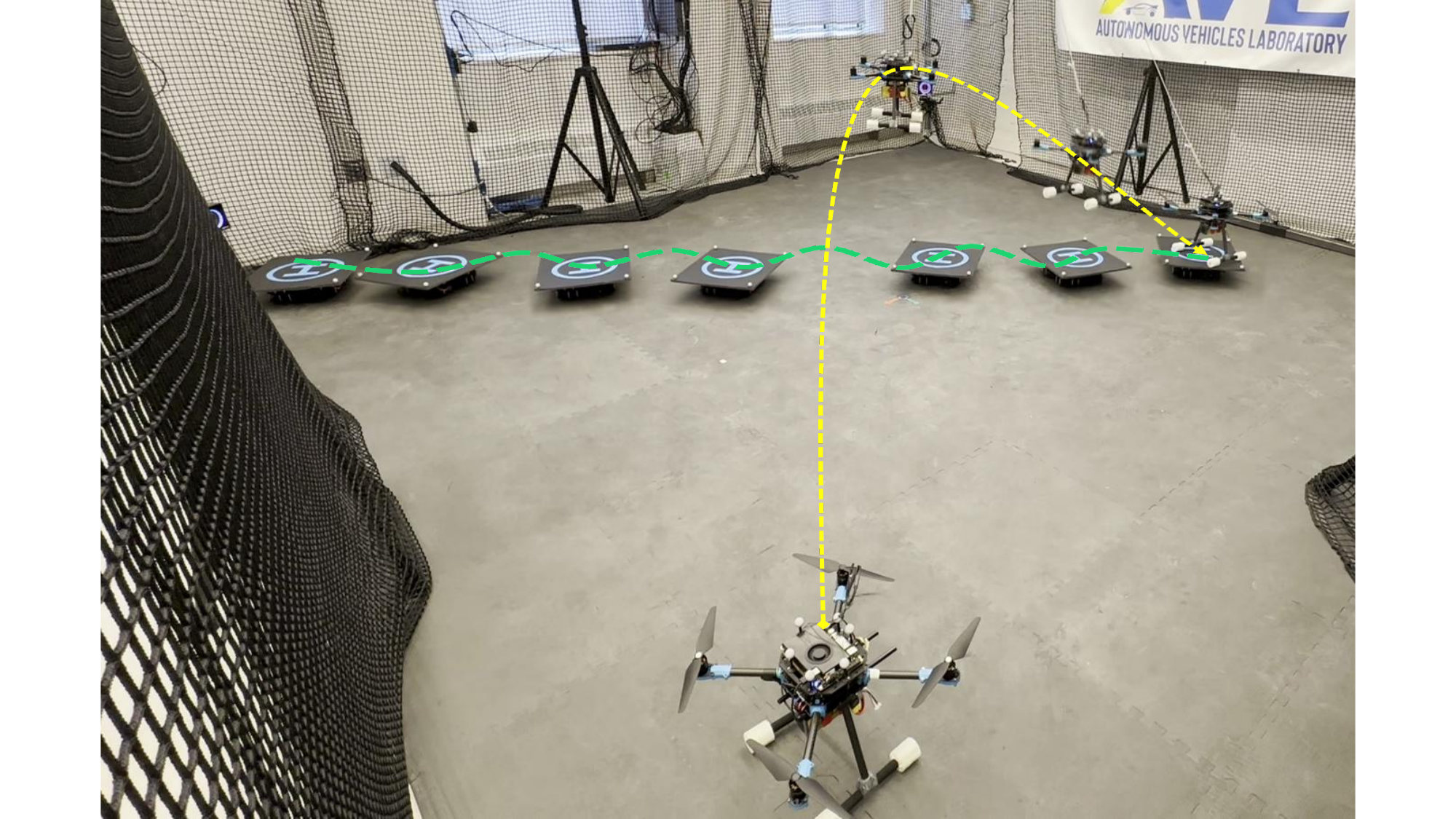}
    \caption{Trajectory of the quadrotor and moving landing platform in a representative dynamic landing trial.}
    \label{fig:dynamic_landing}
\end{figure}

The key challenges of landing on moving platforms, hereafter referred to as dynamic landing (Fig.~\ref{fig:dynamic_landing}), arise in the final stage of landing. During the final descent and touchdown, the vehicle encounters turbulent airflow, including its own wake and platform-induced disturbances, that can lead to flight-path deviations. At the same time, the vehicle must perform rapid position and attitude corrections to synchronize with the platform’s motion for precise landing.

Addressing these challenges requires a careful treatment of state estimation, motion planning, and control. State estimation must not only determine the current states of the vehicle and landing platform but also predict their future states during the final descent. The planner must then generate a feasible agile landing trajectory, while the controller ensures precise tracking of the landing trajectory.

\subsection{Related Work}\label{se:lit_review}
In recent years, several studies have investigated the dynamic landing of MRUAVs \cite{guo2022autonomous, feng2018autonomous, paris2020dynamic, prochazka2024model, serra2016landing, borowczyk2017autonomous, qi2019autonomous, wang2022quadrotor, falanga2017vision, ghommam2017autonomous, gautam2022autonomous, xuan2022quadcopter}. These studies commonly employ a finite state machine or a similar mechanism to transition from conventional navigation to a dedicated multi-stage dynamic landing algorithm, which typically consists of: (i) approaching the vicinity of the landing platform, (ii) tracking the landing pad, and (iii) final descent and touchdown. These three stages are often formulated as a visual servoing problem \cite{serra2016landing}, where an onboard vision-based system informs the vehicle’s relative position and orientation to the landing pad for planning and control purposes.

Several studies rely on fiducial markers, e.g., AprilTag \cite{borowczyk2017autonomous, feng2018autonomous, qi2019autonomous} and ArUco \cite{wang2022quadrotor}, for precise localization. Others incorporate visual-inertial odometry (VIO) to estimate the MRUAV’s state without external positioning infrastructure \cite{falanga2017vision}. Additionally, some approaches integrate optical flow-based velocity estimation \cite{serra2016landing} and ultraviolet direction and ranging systems for enhanced target tracking \cite{prochazka2024model}.

To achieve robust localization of both the MRUAV and the moving platform, many methods employ Kalman filtering, fusing data from inertial measurement units, global navigation satellite systems, and onboard cameras \cite{borowczyk2017autonomous, guo2022autonomous, ghommam2017autonomous, feng2018autonomous, gautam2022autonomous, paris2020dynamic, prochazka2024model, falanga2017vision}.

For landing trajectory planning, several studies employ linear model predictive control (LMPC) to generate optimal trajectories while accounting for system constraints and disturbances \cite{guo2022autonomous, feng2018autonomous, paris2020dynamic, prochazka2024model}. Some works integrate pure pursuit or proportional navigation (PN) guidance for long-range target tracking before landing \cite{borowczyk2017autonomous, gautam2022autonomous}. Polynomial-based trajectory generation has been proposed to ensure smooth and feasible landing trajectories, with certain methods specifically minimizing jerk or snap to optimize control effort and vehicle stability \cite{falanga2017vision}. Other approaches include gradient-based motion planners \cite{wang2022quadrotor} and log-polynomial velocity controllers \cite{gautam2022autonomous}. 

Regarding flight control strategies, a few studies have explored proportional-integral-derivative (PID) controllers \cite{borowczyk2017autonomous,wang2022quadrotor}. \cite{feng2018autonomous} has employed LMPC, though its results are limited to simulations. Nonlinear controllers are more commonly used, including incremental nonlinear dynamic inversion \cite{guo2022autonomous}, boundary-layer sliding mode control \cite{paris2020dynamic}, and other Lyapunov-based designs \cite{serra2016landing, xuan2022quadcopter}. One control strategy that has recently proven effective for improving quadrotor trajectory tracking is nonlinear model predictive control (NMPC), which can leverage both the current and predicted states of the vehicle and landing platform to optimize control actions while accounting for actuator constraints \cite{zhu2023nonlinear, wang2021efficient, nan2022nonlinear, shayan2024nonlinear, sun2022comparative, shayan2025exponential}, along with its variants such as contour NMPC \cite{romero2022model} and linear model bank NMPC \cite{izadi2024multi}.

While previous dynamic landing studies demonstrate significant progress, several critical limitations remain unaddressed. Many existing approaches rely on linear controllers despite the system's inherent nonlinearities and the need for agile maneuvers during the critical touchdown phase \cite{feng2018autonomous, prochazka2024model}. Among studies employing nonlinear controllers \cite{zhu2023nonlinear,guo2022autonomous}, three fundamental challenges persist, as explained below.

First, time-inconsistent landing trajectories remain a critical issue. Existing approaches \cite{guo2022autonomous, feng2018autonomous, paris2020dynamic, prochazka2024model, serra2016landing, borowczyk2017autonomous, qi2019autonomous, wang2022quadrotor, falanga2017vision, ghommam2017autonomous, gautam2022autonomous, xuan2022quadcopter} lack explicit fixed-time touchdown constraints, resulting in unpredictable touchdown timing that increases the risk of landing failures.  

Second, incomplete dynamic modeling limits trajectory generation effectiveness. Several studies such as \cite{gautam2022autonomous, serra2016landing, qi2019autonomous, xuan2022quadcopter} utilize only the landing platform's position for real-time trajectory planning, neglecting velocity information that is crucial for incorporating landing dynamics. While convex optimization-based planning methods \cite{prochazka2024model} offer computational advantages, they frequently employ simplified dynamics models that prove inadequate for the complex nonlinear maneuvers required during final descent. 

Third, most prior works such as \cite{feng2018autonomous,paris2020dynamic, zhu2023nonlinear, borowczyk2017autonomous, qi2019autonomous, falanga2017vision} depend on unscented or extended Kalman filters with fixed noise statistics, which experience significant performance degradation when measurement quality fluctuates due to changing platform distances or environmental conditions.


\subsection{Contributions}
To address the above challenges, we propose an improved dynamic landing algorithm for MRUAVs, validated through hardware experiments. Our main contributions include:  
\begin{enumerate}
    \item Prescribed-time landing trajectory generation: We integrate NMPC with a real-time minimum-jerk planner that enforces a user-specified touchdown time for the terminal descent, targeting consistent timing during dynamic landing.  
    \item Reference feasibility under actuator bounds: We examine the thrust and torque requirements caused by minimum-jerk references and provide sufficient conditions under which the reference caused inputs are within actuator constraints (ensuring feasibility of constraint satisfaction in NMPC tracking). 
    \item Robust adaptive estimation: We use an AUKF that adjusts noise statistics online to account for time-varying measurement quality, and we demonstrate prediction accuracy and repeatable landings in simulation and indoor experiments. 
\end{enumerate}
These contributions address key limitations of previous NMPC-based landing studies such as \cite{zhu2023nonlinear} by enabling prescribed-time touchdown, improving robustness to time-varying noise, and demonstrating consistent experimental performance.

\section{PRELIMINARIES}
\subsection{Notation}
Throughout this paper, unless stated otherwise, we use the following notation standards.
Italic letters indicate scalars, lowercase bold letters represent vectors, and uppercase bold letters denote matrices. 
$\mathbf{I}_n$ indicates $n \times n$ identity matrix.
We use the ENU (East-North-Up) inertial frame \( \mathcal{I} \) and FLU (Forward-Left-Up) body frame \( \mathcal{B} \) as shown in Fig. \ref{fig:frame} . A vector in \( \mathcal{I} \) is denoted as \( {}^{I}\mathbf{p} \), and the rotation from \( \mathcal{B} \) to \( \mathcal{I} \) is represented by a rotation matrix \( {}^I_B\mathbf{R}(\boldsymbol{q}) \) or quaternion \( {}^I_B\boldsymbol{q} = [q_w, q_x, q_y, q_z]^T \in \mathbb{H} \) (Hamilton convention). Quaternion operations include conjugation \( \boldsymbol{q}^* = [q_w, -q_x, -q_y, -q_z]^T \), multiplication \( \otimes \), and vector extraction \( \mathcal{V}(\boldsymbol{q}) = [q_x, q_y, q_z]^T \). \( \mathcal{V}^*(\cdot) \) is used to represent the inverse mapping from a position point, defined as \( \mathcal{V}^*(\mathbf{p}) := [0, \mathbf{p}]^T \), where \( \mathbb{R}^3 \) is mapped to \( \mathbb{H} \).
The rigid-body (pose) transformation in 3D is:
\[
{}^{I}\mathbf{p} = \mathcal{V}\left({}^I_B\boldsymbol{q} \otimes \mathcal{V}^*({}^{B}\mathbf{p}) \otimes {}^I_B\boldsymbol{q}^*\right) + {}^{I}\mathbf{p}_{B_o} = {}^I_B\mathbf{R}(\boldsymbol{q}){}^{B}\mathbf{p} + {}^{I}\mathbf{p}_{B_o},
\]
where \( {}^{I}\mathbf{p}_{B_o} \) is the origin of \( \mathcal{B} \) in \( \mathcal{I} \), and \( \mathbf{R}(\boldsymbol{q}) \) is:
\[
\mathbf{R}(\boldsymbol{q}) = 
\begin{bmatrix}
1 - 2q_y^2 - 2q_z^2 & 2q_x q_y - 2q_w q_z & 2q_x q_z + 2q_w q_y \\
2q_x q_y + 2q_w q_z & 1 - 2q_x^2 - 2q_z^2 & 2q_y q_z - 2q_w q_x \\
2q_x q_z - 2q_w q_y & 2q_y q_z + 2q_w q_x & 1 - 2q_x^2 - 2q_y^2
\end{bmatrix}.
\]

\subsection{Quadrotor Model}
The quadrotor dynamics are derived from six degrees-of-freedom rigid body dynamics. The position $\boldsymbol{\xi}$, velocity $\boldsymbol{\upsilon}$, attitude $\boldsymbol{q}$, and angular velocity $\boldsymbol{\omega}$ dynamics are \cite{li2023nonlinear}
\begin{figure}[t]
    \centering
    \includegraphics[trim={4cm, 9cm, 18cm, 2.9cm}, clip, width = 0.9\linewidth]{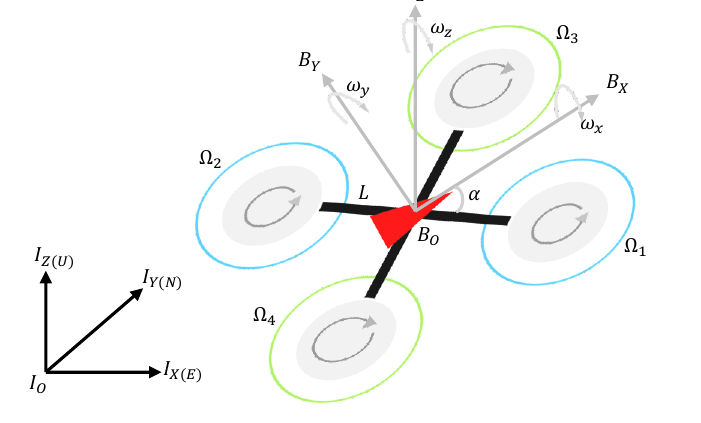}
    \caption{Quadrotor model and coordinate frames}
    \label{fig:frame}
\end{figure}
\begin{equation}
\label{eq:quadrotor_model}
\begin{aligned}
    \left\{
    \begin{array}{l}
        {}^I\dot{\boldsymbol{\xi}} = {}^I\boldsymbol{\upsilon}, \\
        {}^I\dot{\boldsymbol{\upsilon}} = \frac{1}{m} \left( {}^I_B\mathbf{R}(\boldsymbol{q}) \cdot {}^{B}\mathbf{f}_{\mathbf{u}} \right) + {}^{I}\mathbf{g}, \\
        {}^I_B\dot{\boldsymbol{q}} =1/2\cdot {}^I_B\boldsymbol{q} \otimes \mathcal{V}^*({}^{B}\boldsymbol{\omega}), \\
        {}^{B}\dot{\boldsymbol{\omega}} = \mathbf{J}^{-1} \left(-{}^{B}\boldsymbol{\omega} \times \left(\mathbf{J} \cdot {}^{B}\boldsymbol{\omega}\right) + {}^{B}\boldsymbol{\tau}_{\mathbf{u}} \right).
    \end{array}
    \right.
\end{aligned}
\end{equation}
where \( {}^{B}\mathbf{f}_{\mathbf{u}} = [0, 0, f_c]^T \) and \( {}^{B}\boldsymbol{\tau}_{\mathbf{u}} = [\tau_x, \tau_y, \tau_z]^T \) are the control force and torque, \( m \) is the mass, \( \mathbf{J} = \text{diag}(J_{xx}, J_{yy}, J_{zz}) \) is the inertia matrix, and \( {}^{I}\mathbf{g} = [0, 0, -g]^T \) is gravity. The rotor thrust and torque are:
\begin{equation}
 \qquad \qquad   T_i = k_t \Omega_i^2 \quad , \quad \tau_i = k_q \Omega_i^2
\end{equation}
where \( k_t \) and \( k_q \) are thrust and torque coefficients, and \( \Omega_i \) is the rotor speed. Then ${\bf{f}_{\mathbf{u}}}$ and ${\boldsymbol{\tau}_{\mathbf{u}}}$ can be expressed as follows
\begin{IEEEeqnarray}{c}
    \begin{bmatrix} {{f}_{\mathbf{c}}},\boldsymbol{\tau}_{\mathbf{u}} \end{bmatrix}^T
    = \mathbf{G} 
    \begin{bmatrix} T_1,T_2,T_3,T_4 \end{bmatrix}^T,
\end{IEEEeqnarray}
where $\mathbf{G}$ is
\begin{equation}\label{eq:tau}
\mathbf{G} =
\begin{bmatrix}
1 & 1 & 1 & 1 \\
- L \sin \alpha & L \sin \alpha & - L \sin \alpha & L \sin \alpha \\
- L \cos \alpha & L \cos \alpha & - L \cos \alpha & L \cos \alpha \\
- k_q / k_t & - k_q / k_t & k_q / k_t & k_q / k_t
\end{bmatrix},
\end{equation}
with \( L \) as the arm length and \( \alpha \) as the rotor arm angle (Fig. \ref{fig:frame}).

\section{DYNAMIC LANDING ALGORITHM}
\begin{figure*}[ht]
    \centering
    \includegraphics[trim={0cm, 6cm, 0cm, 3cm}, clip, width = 1\linewidth]{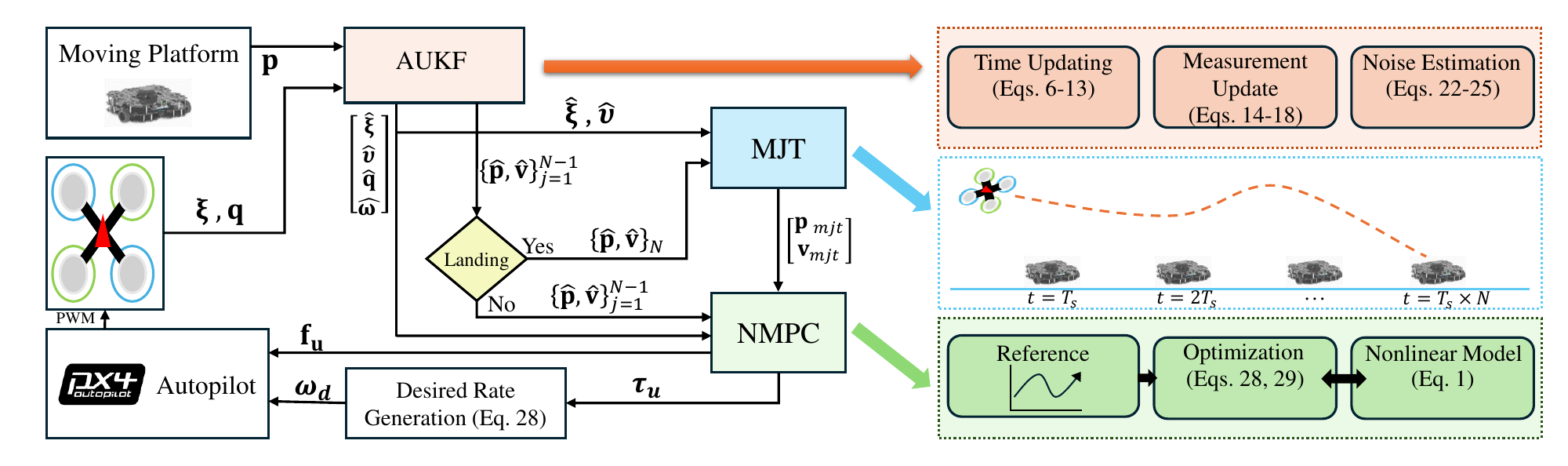}
    \caption{Block diagram of the proposed dynamic landing algorithm}
    \label{fig:block_diagram}
\end{figure*}

Figure \ref{fig:block_diagram} illustrates the overall block diagram of the proposed dynamic landing algorithm. 
We focus on two main modes: (i) tracking, which ensures synchronization of the vehicle with the landing pad, and (ii) final descent and touchdown. 
The initial phases, such as approach to the vicinity of the landing pad, can be handled by conventional navigation methods.

For both tracking and final descent, we employ NMPC as the flight controller. 
A key feature of our framework is the use of an adaptive unscented Kalman filter (AUKF) to estimate the position and velocity of the landing platform over the prediction horizon. 
These estimates are then fed into the NMPC, giving rise to a prescient NMPC formulation \cite{tajeddin2019ecological}. 
This predictive capability is crucial for reliable synchronization, particularly under noisy measurement conditions. 
As shown in Fig.~\ref{fig:aukf}, the AUKF achieves significantly lower velocity prediction errors compared to EKF and UKF, demonstrating its robustness to noise and its critical role in ensuring precise landing performance.

Once the position and velocity of both the vehicle and the landing platform remain bounded within a specified neighborhood for a user-defined duration, the algorithm transitions to the final descent phase. 
At this stage, the MJT planner computes a fixed-time landing trajectory using the vehicle's current states together with the AUKF-predicted future positions and velocities of the platform. 
The use of MJT ensures smooth references, while the fixed-time formulation eliminates the variable descent durations observed in prior works, thereby reducing the likelihood of desynchronization between the vehicle and the platform. 

The landing trajectory is then passed to NMPC for tracking until the quadrotor reaches a predefined proximity to the platform and their relative velocities fall within a specified tolerance, at which point the vehicle’s motors are stopped.

In the following subsections, we provide a detailed description of the AUKF, MJT, and NMPC components.

\subsection{Adaptive Unscented Kalman Filter}
For simultaneous estimation of quadrotor and landing platform states using AUKF, there exist two methods: the joint and the dual filtering methods \cite{aswal2022joint}.
The joint method is more straightforward and easier to implement than the dual method; therefore, it will be our method of choice here.
To this end, introduce an augmented state vector as $\boldsymbol{\zeta} = [ \boldsymbol{\xi}^T, \boldsymbol{q}^T,\boldsymbol{\upsilon}^T, \boldsymbol{\omega}^T, \mathbf{p}^T, \mathbf{v}^T ] $.
The dynamics of the landing platform are expressed by $ \dot{\mathbf{p}} = \mathbf{v}, \quad \dot{\mathbf{v}} = \boldsymbol{\phi}(\mathbf{p}, \mathbf{v})$, where $ \boldsymbol{\phi}(\mathbf{p}, \mathbf{v}) $ represents the motion dynamics of the platform, which may be unknown. 
Therefore, the complete system dynamics and measurement model are given by
\begin{equation}
\boldsymbol{\zeta}_{k+1} = \mathbf{f}(\boldsymbol{\zeta}_k, \boldsymbol{u}_k) + \boldsymbol{w}_k , \:\: \mathbf{y}_{k+1} = \mathbf{h}(\boldsymbol{\zeta}_{k+1}) + \mathbf{v}_k,
\end{equation}
where $\mathbf{f}(\boldsymbol{\zeta_k}, \boldsymbol{u}_k)$ represents the quadrotor and platform dynamics, \(\mathbf{y}_k \) is the measurement vector and  $\mathbf{u}_k$ is controlled input. \(\mathbf{w}_k\) is the process noise with mean \(\mathbf{q}_k\) and covariance \(\mathbf{Q}_k\).
  \(\mathbf{v}_k\) is the measurement noise with mean \(\mathbf{r}_k\) and covariance \(\mathbf{R}_k\). In our implementation, the measurement vector includes the quadrotor position and attitude, as well as the position of the landing platform. Because the velocity of the landing platform is not measured, the unmodeled dynamics of the platform motion for example, accelerations during turns, are modeled by the process noise term in the platform dynamics.

For AUKF predictions, we initialize the state estimate \(\hat{\boldsymbol{\zeta}}_0\) and its covariance \(P_0\), and the noise statistics estimates \(\hat{\mathbf{q}}_0\), \(\hat{\mathbf{Q}}_0\), \(\hat{\mathbf{r}}_0\), and \(\hat{\mathbf{R}}_0\) that are considered known as \(\hat{\boldsymbol{\zeta}}_0 = {E}[\boldsymbol{\zeta_0}]\) and \(\mathbf{P_0} = {E}[(\boldsymbol{\zeta_0-\hat{\boldsymbol{\zeta}}_0})(\boldsymbol{\zeta_0-\hat{\boldsymbol{\zeta}}_0)^T}] \).
This is followed by generating $2n+1$ sigma points \(\chi_{i,k}\) based on the state estimate \(\hat{\boldsymbol{\zeta}}_k\) and covariance \(P_k\)
\begin{equation}
\begin{aligned}
    \boldsymbol{\chi}_{0,k} &= \hat{\boldsymbol{\zeta}}_k, \\
    \boldsymbol{\chi}_{i,k} &= \hat{\boldsymbol{\zeta}}_k + (\sqrt{(n+\lambda)\mathbf{P}_k})_i, \quad i = 1, \dots, n, \\
     \boldsymbol{\chi}_{i+n,k} &= \hat{\boldsymbol{\zeta}}_k - (\sqrt{(n+\lambda)\mathbf{P}_k})_i, \quad i = n+1, \dots, 2n,
    \end{aligned}
\end{equation}    
where $n$ is the number of states and \(\lambda\) is a scaling parameter to minimize the higher-order effects defined as $\lambda = \delta^2 (n + \kappa) - n$. \( \kappa \geq 0 \) must be chosen appropriately to ensure that $\mathbf{P}_\kappa \ge 0$. The parameter \( 0 < \delta \leq 1 \) regulates the spread of the sigma points. Each of these sigma points is propagated through the system dynamics, \( \mathbf{f} \)
\begin{equation}
\boldsymbol{\chi}_{i,k+1|k} = \mathbf{f}(\boldsymbol{\chi}_{i,k}, \mathbf{u}_k) , \quad i = 0,1, \dots, 2n
\end{equation}
The posterior mean and covariance matrix are consequently computed as
\begin{IEEEeqnarray}{rCl}
    \hat{\boldsymbol{\zeta}}_{k+1|k} &=& \sum_{i=0}^{2n} w_i^{(m)} \boldsymbol{\chi}_{i,k+1|k}, \\
    \mathbf{P}_{k+1|k} &=& \sum_{i=0}^{2n} w_i^{(c)} \left( \boldsymbol{\chi}_{i,k+1|k} - \hat{\boldsymbol{\zeta}}_{k+1|k} \right) \nonumber \\
    &&\quad \times \left( \boldsymbol{\chi}_{i,k+1|k} - \hat{\boldsymbol{\zeta}}_{k+1|k} \right)^T + \hat{\mathbf{Q}}_k.
\end{IEEEeqnarray}
The weights \(w_i^{(m)}\) and \(w_i^{(c)}\) are expressed as
\begin{IEEEeqnarray}{rCl}
    w_0^{(m)} &=& \frac{\lambda}{n + \lambda},\:\:  w_0^{(c)} = \frac{\lambda}{n + \lambda} + (1 - \delta^2 + \beta), \\ 
    w_i^{(m)} &=& w_i^{(c)} = \frac{1}{2(n + \kappa)}, \quad i = 1, \dots, 2n,
\end{IEEEeqnarray}
where \(\beta\) is a non-negative parameter, and for a Gaussian prior, it is set to \(\beta = 2\). 

The next step is to propagate each sigma point through the measurement model
\begin{equation}
\boldsymbol{\gamma}_{i,k+1|k} = \mathbf{h}(\boldsymbol{\chi}_{i,k+1|k}) , \quad i = 0,1, \dots, 2n.
\end{equation}
The predicted measurement, innovation covariance, and cross-covariance matrix are calculated as
\begin{IEEEeqnarray}{rCl}
    \hat{\mathbf{y}}_{k+1|k} &=& \sum_{i=0}^{2n} w_i^{(m)} \boldsymbol{\gamma}_{i,k+1|k},  \\ 
    \mathbf{P}_{yy,k+1|k} &=& \sum_{i=0}^{2n} w_i^{(c)} [\boldsymbol{\gamma}_{i,k+1|k} - \hat{\mathbf{y}}_{k+1|k}] \notag \\
    && \quad \times [\boldsymbol{\gamma}_{i,k+1|k} - \hat{\mathbf{y}}_{k+1|k}]^T + \hat{\mathbf{R}}_k, \\ 
    \mathbf{P}_{xy,k+1|k} &=& \sum_{i=0}^{2n} w_i^{(c)} [\boldsymbol{\chi}_{i,k+1|k} - \hat{\boldsymbol{\zeta}}_{k+1|k}] \notag \\
    && \quad \times [\mathbf{\boldsymbol{\gamma}}_{i,k+1|k} - \hat{\mathbf{y}}_{k+1|k}]^T
\end{IEEEeqnarray}
Using these matrices, we can compute the Kalman filter gain and update the state estimate and covariance matrix as follows
\begin{IEEEeqnarray}{rCl}
    \mathbf{K}_{k+1}  &=& \mathbf{P}_{xy,k+1|k} \mathbf{P}_{yy,k+1|k}^{-1},  \\ 
    \hat{\boldsymbol{\zeta}}_{k+1} &=& \hat{\boldsymbol{\zeta}}_{k+1|k} + \mathbf{K}_{k+1} (\mathbf{y}_{k+1} - \hat{\mathbf{y}}_{k+1|k}),\\ 
    \mathbf{P}_{k+1} &=& \mathbf{P}_{k+1|k} - \mathbf{K}_{k+1} (\mathbf{P}_{yy,k+1|k} +\epsilon \mathbf{I} ) \mathbf{K}_{k+1}^T
    \label{eq:cov_update}
\end{IEEEeqnarray}

Note that we added a regularization term $\epsilon \mathbf{I}_{19}$ in \eqref{eq:cov_update} where $\epsilon$ is a small number ($\epsilon=10^{-6}$).
This is to prevent singularity and numerical instability issues that may arise in updating the $\mathbf{P}_{k+1}$.

As mentioned earlier, the primary challenge with standard UKF is that it assumes that \(\mathbf{Q}_k\) and \(\mathbf{R}_k\) are known and time-invariant. 
With AUKF, we aim to dynamically estimate the noise statistics (\( \hat{\mathbf{Q}}, \hat{\mathbf{q}}, \hat{\mathbf{R}}, \hat{\mathbf{r}} \)) that optimize the posterior density function based on measurements \( \mathbf{y}_k \) and states \( \boldsymbol{\zeta}_k \)  by formulating the estimation problem using Maximum A Posteriori (MAP) estimation \cite{zhao2009adaptive}. This approach optimizes the posterior density function based on measurements  $\mathbf{y}_k$  and states $\boldsymbol{\zeta}_k$, as follows

\begin{IEEEeqnarray}{rCl}
    J^* &=& p(\boldsymbol{\mathcal{Z}}(k), \mathbf{q}, \mathbf{Q}, \mathbf{r}, \mathbf{R} | \mathbf{Y}(k))  \nonumber \\
        &\propto& p(\mathbf{Y}(k)|\boldsymbol{\mathcal{Z}}(k), \mathbf{q}, \mathbf{Q}, \mathbf{r}, \mathbf{R})  \nonumber \\
        && \times p(\boldsymbol{\mathcal{Z}}(k)|\mathbf{q}, \mathbf{Q}, \mathbf{r}, \mathbf{R}).
\end{IEEEeqnarray}

Where $\mathcal{Z}(k)$ denotes the measurement vector at time step $k$. Assuming Gaussian noise distributions, the state and measurement likelihoods are
\begin{IEEEeqnarray}{l}
    \nonumber 
    p(\boldsymbol{\mathcal{Z}}(k)|\mathbf{q}, \mathbf{Q}) \propto  \\ 
    \quad \prod_{j=1}^{k} 
    \exp \left( -\frac{1}{2}  
    \left\| \boldsymbol{\zeta}_j - \boldsymbol{\mathcal{F}}_{j-1}(\boldsymbol{\zeta}_{j-1}) - \mathbf{q} 
    \right\|_{\mathbf{Q}^{-1}}^2 \right),
\end{IEEEeqnarray}
\begin{IEEEeqnarray}{l}
    \nonumber
    p(\mathbf{Y}(k)|\boldsymbol{\mathcal{Z}}(k), \mathbf{r}, \mathbf{R}) \propto  \\  
    \quad \prod_{j=1}^{k} 
    \exp \left( -\frac{1}{2}  
    \left\| \mathbf{y}_j - \boldsymbol{\mathcal{H}}_j(\boldsymbol{\zeta}_j) - \mathbf{r} 
    \right\|_{\mathbf{R}^{-1}}^2 \right).
\end{IEEEeqnarray}

Maximizing the posterior probability and solving for the optimal noise statistics yield \cite{xing2011adaptive}
\begin{IEEEeqnarray}{rCl}
    \hat{\mathbf{q}}_k &=& \frac{1}{k} \sum_{j=1}^{k} [\hat{\boldsymbol{\zeta}}_j - \boldsymbol{\mathcal{F}}_{j-1}(\hat{\boldsymbol{\zeta}}_{j-1})],
\end{IEEEeqnarray}
\begin{IEEEeqnarray}{rCl}
    \hat{\mathbf{Q}}_k &=& \frac{1}{k} \sum_{j=1}^{k} [\hat{\boldsymbol{\zeta}}_j - \boldsymbol{\mathcal{F}}_{j-1}(\hat{\boldsymbol{\zeta}}_{j-1}) - \hat{\mathbf{q}}] \nonumber \\
    && \quad \times [\hat{\boldsymbol{\zeta}}_j - \boldsymbol{\mathcal{F}}_{j-1}(\hat{\boldsymbol{\zeta}}_{j-1}) - \hat{\mathbf{q}}]^T, 
\end{IEEEeqnarray}
\begin{IEEEeqnarray}{rCl}
    \hat{\mathbf{r}}_k = \frac{1}{k} \sum_{j=1}^{k} [\mathbf{y}_j - \boldsymbol{\mathcal{H}}_j(\hat{\boldsymbol{\zeta}}_j|j-1)], 
\end{IEEEeqnarray}
\begin{IEEEeqnarray}{rCl}
    \hat{\mathbf{R}}_k &=& \frac{1}{k} \sum_{j=1}^{k} [\mathbf{y}_j - \boldsymbol{\mathcal{H}}_j(\hat{\boldsymbol{\zeta}}_j|j-1) - \hat{\mathbf{r}}] \nonumber \\
    && \quad \times [\mathbf{y}_j - \boldsymbol{\mathcal{H}}_j(\hat{\boldsymbol{\zeta}}_j|j-1) - \hat{\mathbf{r}}]^T.
\end{IEEEeqnarray}
The system functions \( \boldsymbol{\mathcal{F}} \) and \( \boldsymbol{\mathcal{H}} \) are computed as follows
\begin{align}
    \boldsymbol{\mathcal{F}}_{j-1}(\hat{\boldsymbol{\zeta}}_{j-1}) &= \sum_{i=0}^{2n} w_i^{(m)} \boldsymbol{\mathcal{F}}(\boldsymbol{\chi}_{i,j-1|j-1}), \\
    \boldsymbol{\mathcal{H}}_j(\hat{\boldsymbol{\zeta}}_j|j-1) &= \sum_{i=0}^{2n} w_i^{(m)} \boldsymbol{\mathcal{H}}(\boldsymbol{\chi}_{i,j|j-1}).
\end{align}

\subsection{Fixed-Time MJT Planning}

To eliminate the timing inconsistency inherent in conventional NMPC-based landing \cite{zhu2023nonlinear}, we generate fixed-time minimum-jerk landing trajectory in real-time. Given the current quadrotor state and the predicted motion of the landing platform (provided by the AUKF), we compute smooth reference trajectories that guarantee convergence at a prescribed touchdown time $T_f$. This formulation ensures predictable synchronization with the platform while minimizing control effort.

The MJT generation problem is posed as a variational optimization:

\begin{equation}
\min_{\mathbf{p}(\cdot)} \; J = \tfrac{1}{2} \int_{0}^{T_f} \left\| \tfrac{d^3\mathbf{p}(t)}{dt^3} \right\|^2_{W_j} \, dt
\label{eq:mjt}
\end{equation}
subject to the boundary conditions $\mathbf{p}(0)=\mathbf{p}_0, \, \dot{\mathbf{p}}(0)=\mathbf{v}_0, \, \ddot{\mathbf{p}}(0)=\mathbf{a}_0$ and 
$\mathbf{p}(T_f)=\mathbf{p}_T, \, \dot{\mathbf{p}}(T_f)=\mathbf{v}_T, \, \ddot{\mathbf{p}}(T_f)=\mathbf{a}_T$. The optimal solution corresponds to a fifth-order polynomial trajectory whose coefficients are obtained by solving the associated boundary value problem. At each time step, the MJT planner generates synchronized position and velocity setpoints $\{\mathbf{p}_{mjt}(t), \mathbf{v}_{mjt}(t)\}$ that are tracked by the NMPC. This guarantees fixed-time convergence and avoids the variable landing durations observed in prior NMPC landing studies. For brevity, the full derivation of the polynomial coefficients is omitted, as it follows standard MJT planning formulations \cite{piazzi2000global}. In our implementation, the touchdown time $T_f$ is chosen as a fixed constant for the terminal descent.

\subsection{Controller Design}
Let us begin by defining the state vector ${\bf s} = \left[{\boldsymbol{\hat{\xi}}}^T,{\boldsymbol{\hat{q}}}^T,{\boldsymbol{\hat{\upsilon}}}^T,{\boldsymbol{\hat{\omega}}}^T\right]^T$ provided by AUKF. The tracking error is defined as $\mathbf{e}= \mathbf{s} - \mathbf{s}_r$, where $\mathbf{s}_r = \left[{\boldsymbol{\xi}}_r^T,{\boldsymbol{q}}_r^T,{\boldsymbol{\upsilon}}_r^T,{\boldsymbol{\omega}}_r^T\right]^T$ represents the reference trajectory. ${\boldsymbol{q}}_r = \left[ 1, 0, 0, 0 \right]^T$, and ${\boldsymbol{\omega}}_r = \left[ 0, 0, 0 \right]^T$. Since the platform attitude remains approximately level in our experiments, we use a level attitude reference during landing.  $\boldsymbol{\xi}_r$, and $\boldsymbol{\upsilon}_r$, take different values depending on the mission stage. During the tracking phase, they correspond to the estimated position and velocity of the platform, denoted as $\hat{\mathbf{p}}$ and $\hat{\mathbf{v}}$ respectively. In the landing phase, they are set to MJT $\mathbf{p}_{mjt}$ and $\mathbf{v}_{mjt}$.

The goal is to determine the control signal $\mathbf{u}= \left[{}^{B}\mathbf{f}_{\mathbf{u}}, {}^{B}\boldsymbol{\tau}_{\mathbf{u}} \right]^T$ such that ${\bf s}$ reaches the desired trajectory ${\bf s}_r$. To this end, we adopt the standard NMPC formulation 
\begin{equation}
\label{eq:nmpcCost}
 \qquad  \quad   \min_{\mathbf{u}_k}  \left(\mathcal{M}\left(\mathbf{e}_N\right) + \sum_{k=0}^{N-1} \mathcal{L}\left(\mathbf{e}_k, \mathbf{u}_k\right) \right),
\end{equation}
subject to
\begin{equation}
\label{eq:nmpcConstraints}
\begin{array}{c}
 \qquad  \quad  \mathbf{s}_{k+1} = \mathbf{g}\left(\mathbf{s}_k,\mathbf{u}_k\right), \; \mathbf{s}_0 = \mathbf{s}_{init}, \\
\mathbf{s}_k \in \mathcal{S}, \; \mathbf{u}_k \in \mathcal{U},
\end{array}
\end{equation}
where $\mathcal{L}\left(\mathbf{e}_k, \mathbf{u}_k\right)$ is the running cost, $\mathcal{M}\left(\mathbf{e}_N\right)$ is the terminal cost, $\mathbf{s}_{k+1} = \mathbf{g}\left(\mathbf{s}_k,\mathbf{u}_k\right)$ is the discretized nonlinear dynamics \eqref{eq:quadrotor_model}, $\mathcal{S}$ and $\mathcal{U}$ are the sets of allowable $\bf{s}$ and ${\bf{u}}$, and $N$ is the prediction horizon. For our implementations, we set
\begin{equation}\label{eq:cost_function}
\begin{array}{c}
\qquad  \quad  \mathcal{L}(\mathbf{e}_k, \mathbf{u}_k) = \mathbf{e}_k^T \mathbf{Q}_\mathbf{e} \mathbf{e}_k + \mathbf{u}_k^T \mathbf{Q}_\mathbf{u} \mathbf{u}_k, \\ 
\mathcal{M}(\mathbf{e}_N) = \mathbf{e}_N^T \mathbf{Q}_\mathbf{e} \mathbf{e}_N.
\end{array}
\end{equation}
where $ {\bf Q_e}  > 0$ and ${\bf Q}_{{\bf{u}}} > 0$ are the weight matrices. 

After obtaining the control input by solving \eqref{eq:nmpcCost}, the desired body rates $\boldsymbol{\omega_d}$ are computed based on $\boldsymbol{\tau}_u$ by
\begin{equation}
\label{eq:desired_rate}
\begin{array}{c}
 \qquad  \quad  \dot{\boldsymbol{\omega}}_d = \mathbf{J}^{-1} \left(\boldsymbol{\tau_u} - \boldsymbol{\omega} \times \mathbf{J} \boldsymbol{\omega}\right), \\
\boldsymbol{\omega}_d = \boldsymbol{\omega} + \dot{\boldsymbol{\omega}}_d \cdot T_s.
\end{array}
\end{equation}
where $T_s$ is the sample time.
\subsection{Theoretical Analysis of MJT Integration with NMPC}
\label{theory}
We analyze how the MJT reference influences the feasibility of NMPC under actuator limits.
Consider the quadrotor dynamics in \eqref{eq:quadrotor_model} with control inputs
${}^{B}\mathbf{f}_{\mathbf{u}}=[0,0,f_c]^\top$ and 
${}^{B}\boldsymbol{\tau}_{\mathbf{u}}=[\tau_x,\tau_y,\tau_z]^\top$. 
Let the MJT reference $(\mathbf{p}_{\mathrm{mjt}},\dot{\mathbf{p}}_{\mathrm{mjt}},\ddot{\mathbf{p}}_{\mathrm{mjt}})$ be used to define
the desired translational acceleration
\[
\mathbf{a}_{\mathrm{ref}}(t):=\ddot{\mathbf{p}}_{\mathrm{mjt}}(t), 
\qquad
\|\mathbf{a}_{\mathrm{ref}}\|_\infty \le \alpha_a,
\qquad
\|\dot{\mathbf{a}}_{\mathrm{ref}}\|_\infty \le \alpha_j.
\]
where $\alpha_a>0$ and $\alpha_j>0$ denotes a uniform bound on the acceleration, and jerk respectively. 
Define $\mathbf{a}_g(t):=\mathbf{a}_{\mathrm{ref}}(t)-{}^{I}\mathbf{g}$ and 
the desired thrust direction ${}^{I}\mathbf{b}_{3,d}(t):=\mathbf{a}_g(t)/\|\mathbf{a}_g(t)\|$. 
The corresponding thrust magnitude is
\begin{equation}
    f_{c,d}(t) = m\,\|\mathbf{a}_{g}(t)\|.
    \label{eq:fc_desired_polished}
\end{equation}
By the triangle inequality, $f_{c,d}(t)$ is uniformly bounded as
\begin{equation}
    0 \;\le\; f_{c,d}(t) \;\le\; m\big(\alpha_a+\|{}^{I}\mathbf{g}\|\big) \;=:\; \bar f_c,
    \qquad \forall t\in[0,T_f].
    \label{eq:fc_bound_polished}
\end{equation}
where $\bar f_c$ is the maximum thrust demand induced by MJT references.
Next, suppose the attitude controller tracks ${}^{I}\mathbf{b}_{3,d}$ with bounded angular velocity and acceleration, i.e.,
\begin{equation}
\|{}^{B}\boldsymbol{\omega}(t)\|\le \bar\omega, 
\qquad 
\|\dot{{}^{B}\boldsymbol{\omega}}(t)\|\le \kappa_\omega^0+\kappa_\omega^1\,\alpha_a+\kappa_\omega^2\,\alpha_j,
\label{eq:omega_bounds_polished}
\end{equation}
where $\bar\omega$ is a uniform bound on body angular velocity magnitude, and the constants $\kappa_\omega^0,\kappa_\omega^1,\kappa_\omega^2>0$ capture controller-dependent gains and geometry that scale with the acceleration and jerk bounds. From the rotational dynamics in \eqref{eq:quadrotor_model}, the control torques satisfy

{\small
\begin{equation}
\|{}^{B}\boldsymbol{\tau}_{\mathbf{u}}(t)\|
\le
\underbrace{\|{}^{B}\boldsymbol{\omega}(t)\times\mathbf{J}{}^{B}\boldsymbol{\omega}(t)\|}_{\le \|\mathbf{J}\|\bar\omega^2}
+
\underbrace{\|\mathbf{J}\|\;\|\dot{{}^{B}\boldsymbol{\omega}}(t)\|}_{\le \|\mathbf{J}\|(\kappa_\omega^0+\kappa_\omega^1\alpha_a+\kappa_\omega^2\alpha_j)}
=: \bar\tau,
\label{eq:tau_bound_polished}
\end{equation}
}
so that componentwise
\[
|f_c|\le \bar f_c,
\qquad
|\tau_i|\le \bar\tau_i \quad (i\in\{x,y,z\}),
\]
with $\bar\tau_i$ induced by $\bar\tau$ and the allocation matrix $\mathbf{G}$. Note that the Coriolis contribution is bounded by $\bar\tau_{\text{cori}} \le \|\mathbf{J}\|\bar\omega^2$.

Finally, assume the actuator constraints are
\[
\begin{aligned}
T_i \in [T_{\min},T_{\max}], \quad i=1,\dots,4, \\
\Longleftrightarrow \quad f_c \in [\underline f_c,\overline f_c], \quad \boldsymbol{\tau}_{\mathbf{u}} \in \mathcal{B}_\infty(\overline{\boldsymbol{\tau}})
\end{aligned}
\]
where the thrust interval and torque box $\mathcal{B}_\infty(\overline{\boldsymbol{\tau}})$ are induced by the allocation matrix $\mathbf{G}$ in \eqref{eq:tau}. If the bounds satisfy
\begin{equation}
\bar f_c \le \overline f_c - \gamma_f,
\qquad
\bar\tau_i \le \overline\tau_i - \gamma_{\tau,i},
\quad i\in\{x,y,z\},
\label{eq:margin_conditions_polished}
\end{equation}
for some positive margins $\gamma_f$ and $\gamma_{\tau,i}$, then the tightened input set
\[
\mathcal{U}_\gamma := \Big\{(f_c,\boldsymbol{\tau}_{\mathbf{u}})\;:\;
f_c \in [\underline f_c+\gamma_f,\;\overline f_c-\gamma_f],\;\;
|\tau_i| \le \overline\tau_i-\gamma_{\tau,i} \Big\}
\]
is feasible for the entire MJT reference horizon. Consequently, NMPC admits a strictly positive constraint margin
\[
\gamma = \min\{\gamma_f,\gamma_{\tau,x},\gamma_{\tau,y},\gamma_{\tau,z}\},
\]
which guarantees that actuator limits are respected throughout the maneuver.

The above results indicate that the thrust and torque demands induced by the MJT reference are uniformly bounded, and a nonempty tightened input set $\mathcal{U}_\gamma$ is guaranteed to exist when the actuator margins are satisfied. This is a sufficient condition that lends support to the satisfaction of constraints in NMPC tracking. In this case, the bounded acceleration and jerk values lead to less aggressive input transients, which in turn ensures smooth control allocation during the terminal descent phase. We note that this result addresses reference-induced inputboundedness and does not, by itself, guarantee NMPC convergence or optimality.

\section{SIMULATION AND HARDWARE EXPERIMENTS}
\subsection{Simulations}
\begin{figure}[t]
    \centering
    \includegraphics[trim={5cm, 2cm, 2cm, 2cm}, clip, width = 1\linewidth]{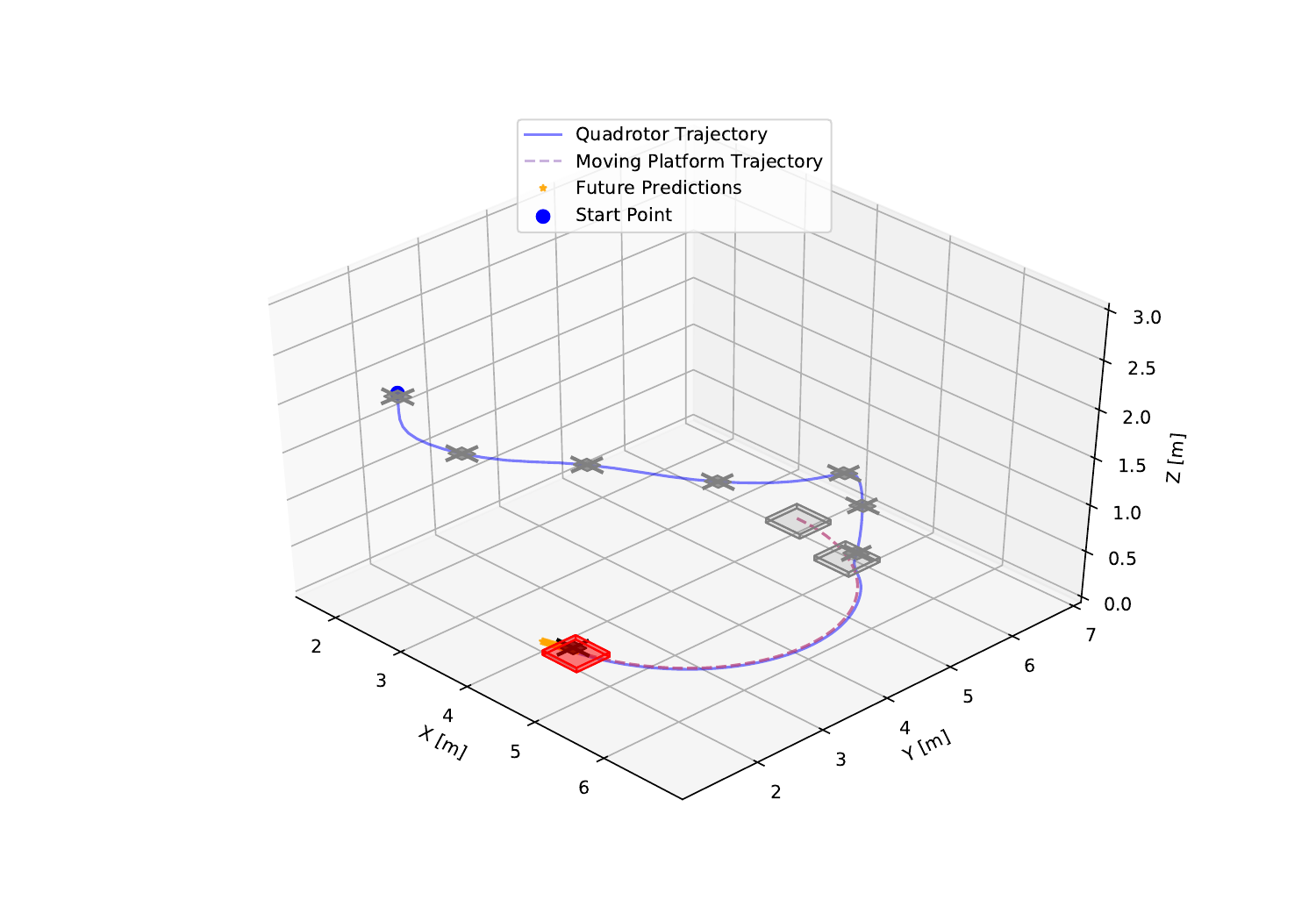}
    \caption{Simulated landing of a quadrotor on a moving platform.}
    \label{fig:sim}
\end{figure}

We conducted numerical simulations to evaluate the algorithm prior to hardware implementations. We carried out simulation assuming that only the position and orientation of the vehicle and the position of the landing platform is available.
The NMPC problem is solved in real time using acados with the partial-condensing HPIPM QP solver \cite{verschueren2022acados}, running at 100 Hz.
We set the initial position of the quadrotor and the landing platform as $(2, 2, 2)^T$ and $(4, 6, 0.1)^T$, respectively. The quadrotor begins tracking the landing platform and, once within a sufficient proximity, initiates the landing maneuver. The results of the simulated experiments are illustrated in Fig. \ref{fig:sim}.


The parameters used in the simulations are set as follows: 
$m = 2.3$~kg, $L = 0.25$~m, $g = 9.8$~m/s\textsuperscript{2}, 
$\alpha = 45^\circ$, and inertia 
$\mathbf{J} = \mathrm{diag}(0.03,\,0.03,\,0.06)$~kg$\cdot$m\textsuperscript{2}. 
For the NMPC, the weights are chosen as 
$\mathbf{Q_e} = 10^3\mathrm{diag}(8,\,8,\,8,\,6,\,4,\,4.5,\,1,\,1,\,1,\,1,\,4,\,4,\,4)$ 
and $\mathbf{Q_u} = 100\mathbf{I}_3$, with a sampling time $T_s = 0.01$~s 
and horizon length $N = 80$. 
The AUKF parameters are set as $\delta = 0.15$, $\beta = 2$, $\kappa = 0$, and 
$n = 19$, with initial covariance $\mathbf{P}_0 = 0.1\mathbf{I}_{19}$. 
The measurement noise covariance is 
$\hat{\mathbf{R}}_0 = \mathrm{diag}(0.01,\,0.01,\,0.01,\,0.2,\,0.2,\,0.2)$, 
and the process noise covariance is 
$\hat{\mathbf{Q}}_0 = \mathrm{diag}(0.01\mathbf{I}_3,\,0.01\mathbf{I}_3,\,0.001\mathbf{I}_4,\,0.05\mathbf{I}_3,\,0.01\mathbf{I}_3,\,0.2\mathbf{I}_3)$.

\subsection{Hardware Experiments}
For hardware experiments, we utilized a modified Holybro X500 equipped with a Pixhawk flight controller and a companion computer featuring an Intel Core i5-1340P processor. 
We implement NMPC within the Robot Operating System (ROS) framework on the onboard companion computer, with its outputs sent to PX4 Autopilot via the MAVROS protocol at $100$ Hz.
We used a TurtleBot3 Waffle Pi to carry the landing platform. The landing pad measured 50 [cm] × 50 [cm], mounted the ground robot as shown in Fig. \ref{fig:dynamic_landing}.
We conducted the experiments in an indoor environment with an OptiTrack motion capture system that measured the position of the vehicle and the landing platfrom. 
Since the motion capture position measurements are accurate with negligible noise, Gaussian noise $\mathcal{N}(0,0.01)$~m\textsuperscript{2} was added to the OptiTrack position data to emulate realistic sensing conditions.

\begin{figure}[t]
    \centering
    \includegraphics[trim={3cm, 8.5cm, 3cm, 8.5cm}, clip, width = \linewidth]{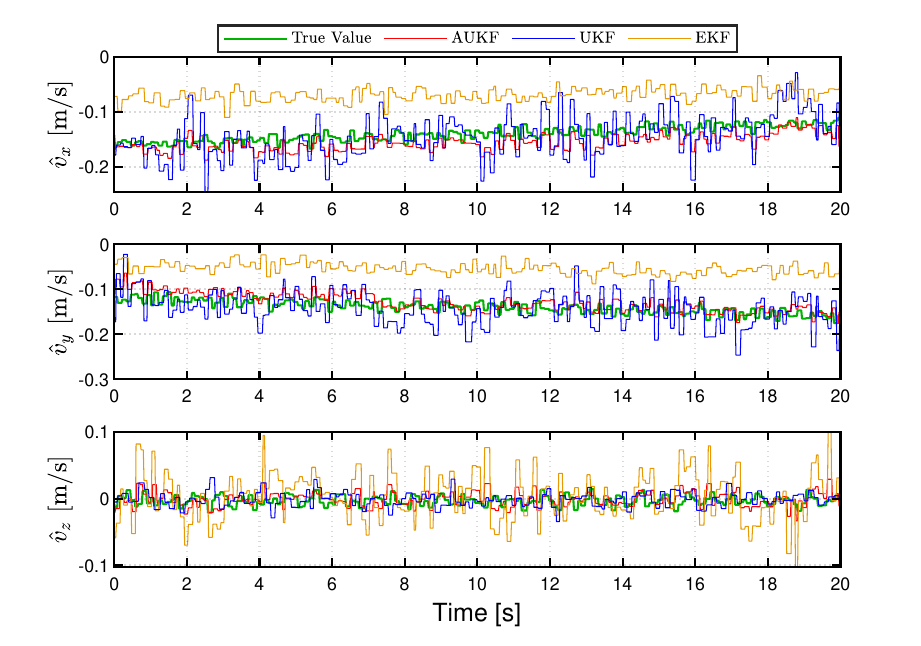}
    \caption{Velocity prediction error for AUKF, UKF, and EKF}
    \label{fig:aukf}
\end{figure}

To evaluate the performance of the AUKF, we compared its performed with EKF and the standard UKF.
Figure \ref{fig:aukf} presents the estimated velocity of the landing platform for the three estimators, followed by the comparison of their root meant square error (RMSE) in Tab. \ref{tab:vel_rmse}.
It is evident that AUKF maintains much lower prediction error compared to the UKF and EKF (on average, 29.66\% lower than UKF and 63.26\% lower than EKF).
This lower prediction error directly translates into improved landing precision.
\begin{table}[t]
    \centering
    \caption{RMSE of velocity predictions for AUKF, UKF, and EKF}
    \label{tab:vel_rmse}
    \begin{tabular}{lccc}
        \toprule
        \textbf{Velocity Component} & \textbf{AUKF} [m/s] & \textbf{UKF} [m/s] & \textbf{EKF} [m/s] \\
        \midrule
        $\hat{v}_x$ & 0.025 & 0.038 & 0.067 \\
        $\hat{v}_y$ & 0.026 & 0.039 & 0.096 \\
        $\hat{v}_z$ & 0.022 & 0.028 & 0.048 \\
        \bottomrule
    \end{tabular}
\end{table}

\begin{figure}[t]
    \centering
    \includegraphics[trim={4cm, 8cm, 4cm, 8cm}, clip, width = \linewidth]{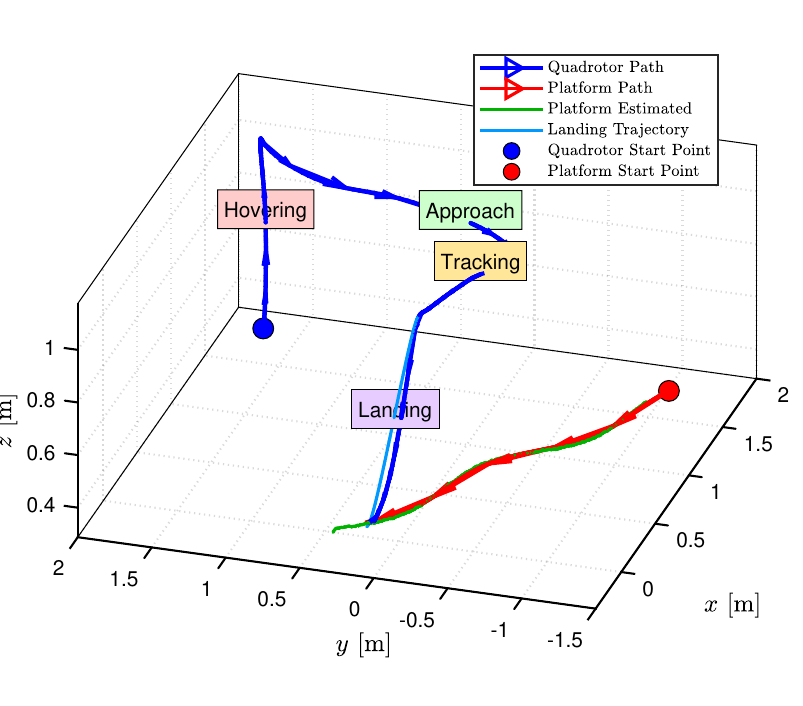}
    \caption{3D view of quadrotor and landing platform trajectories across different mission states in a dynamic landing trial.}
    \label{fig:3D}
\end{figure}

Figure \ref{fig:3D} illustrates the vehicle and landing platform trajectory over a dynamic landing trial.
The landing platform follows a piecewise-smooth trajectory, and the motion pattern is varied across trials.
The vehicle ascends from the initial position $(1.8,1.8,0.34)$ hovers for $2$ [s]. Next, given the landing pad position, the vehicle approaches the landing platform and synchronizes its motion with the platform.
The final descent starts at $t=8.5$ [s] takes approximately $1.5$ [s], terminating with a successful landing on the platform.
The accuracy of the AUKF’s predictions of the landing platform’s position is evident in this figure. Additionally, the computed landing MJT demonstrates a rapid descent and how closely NMPC has followed the planned trajectory.

\begin{figure}[t]
    \centering
    \includegraphics[trim={3cm, 8.5cm, 3cm, 8.5cm}, clip, width = \linewidth]{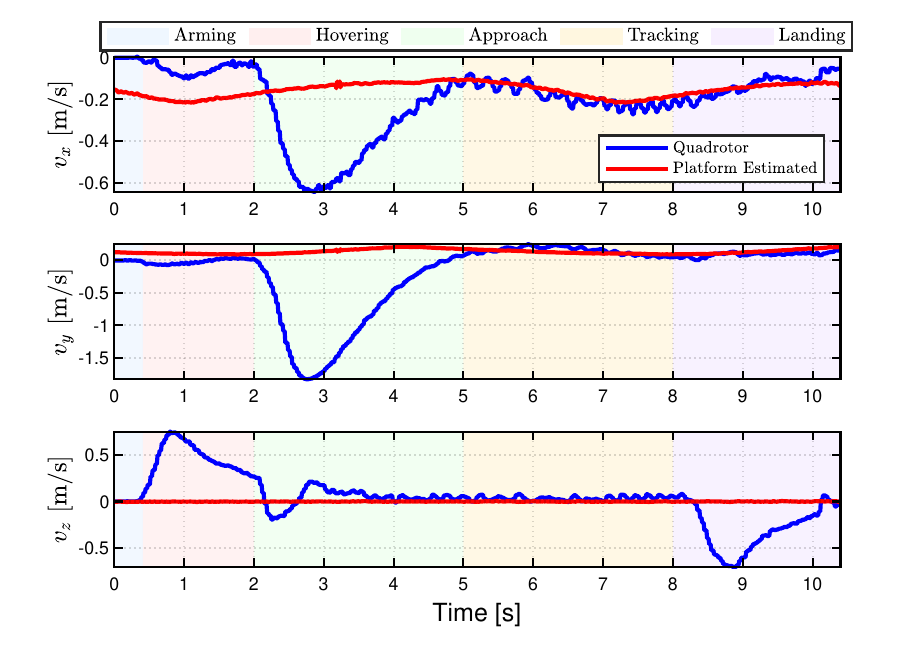}
    \caption{Quadrotor velocity and landing platform estimated velocity across multiple stages of a dynamic landing trial.}
    \label{fig:velocity}
\end{figure}

Figure \ref{fig:velocity} illustrates the velocity of the quadrotor alongside the estimated velocity of the landing platform, as computed by the AUKF. 
The quadrotor’s velocity closely follows that of the platform throughout the tracking phase, demonstrating effective synchronization.

\begin{figure}[t]
    \centering
    \includegraphics[trim={3cm, 8.5cm, 3cm, 8.5cm}, clip, width = \linewidth]{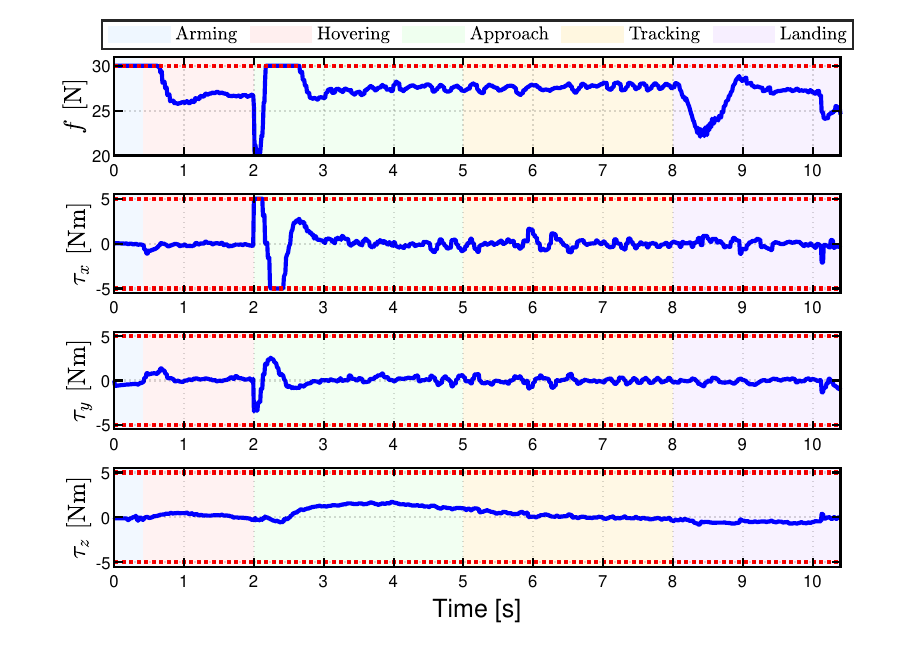}
    \caption{Thrust and moments prescribed by the NMPC across multiple stages of a dynamic landing trial.}
    \label{fig:force}
\end{figure}

Figure \ref{fig:force} depicts the thrust and moments prescribed by NMPC alongside the actuator constraints. Except for the initial approach phase, which required the vehicle to rapidly reach the landing platform, the commanded thrust and moments remain well within the input constraints. This observation is consistent with our theoretical analysis, which shows that minimum-jerk references induce uniformly bounded thrust and torque demands, which is consistent with the reference feasibility analysis in  \ref{theory} and supports constraint satisfaction during landing.


\begin{figure}[t]
    \centering
    \includegraphics[trim={3cm, 6.5cm, 3cm, 6cm}, clip, width = 0.75\linewidth]{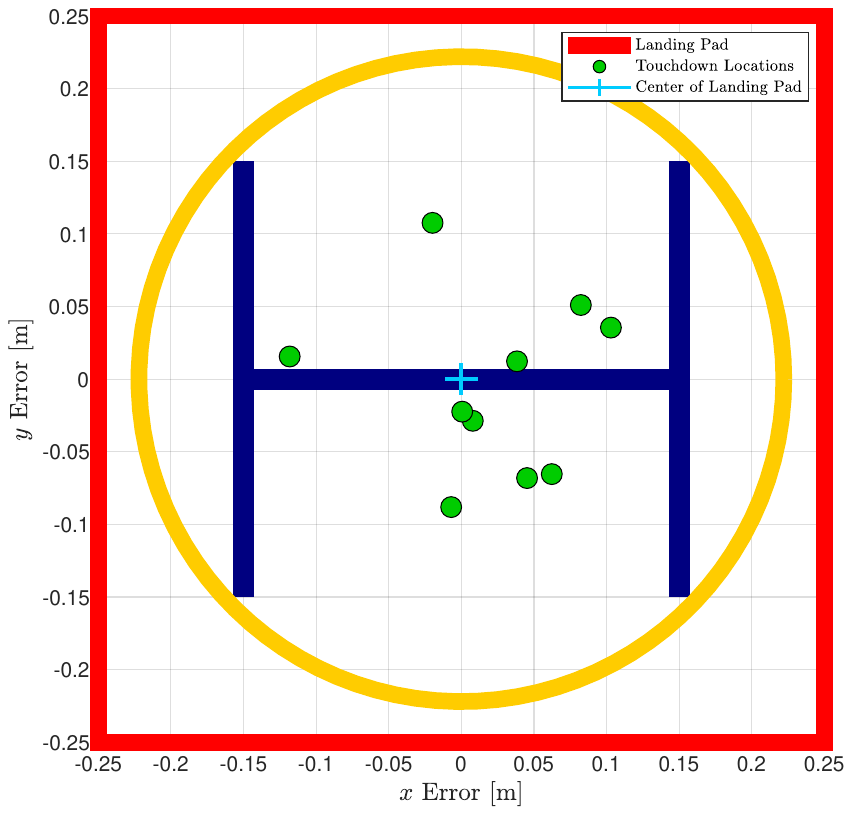}
    \caption{Quadrotor touchdown position for 10 dynamic landing trials.}
    \label{fig:accuracy}
\end{figure}

\begin{figure}[t]
    \centering
    \includegraphics[trim={3cm, 10cm, 3cm, 10cm}, clip, width = \linewidth]{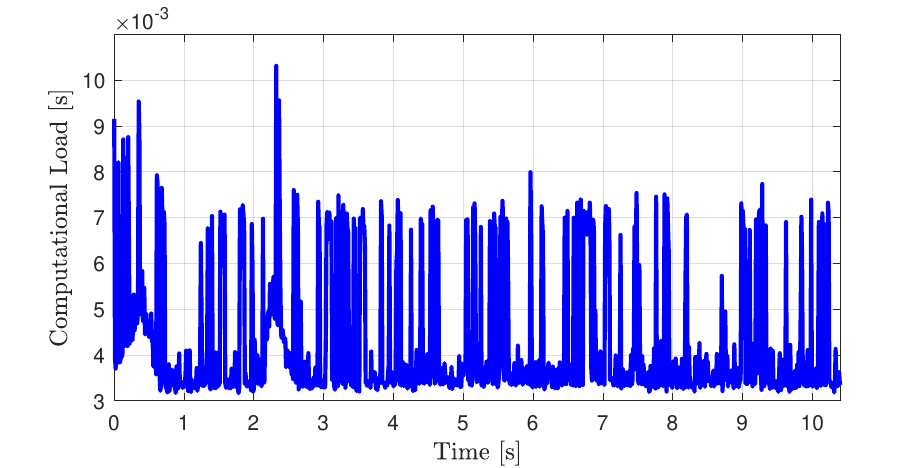}
    \caption{Computational load of the NMPC solver.}
    \label{fig:solver}
\end{figure}

To assess the repeatability of the results, we conducted 10 dynamic landing trials, each starting from different initial conditions and with varying motion patterns for the landing platform. All 10 trials resulted in successful dynamic landings.
Figure \ref{fig:accuracy} illustrates the final touchdown positions of the quadrotor on the landing pad. The mean Euclidean distance from the center of the landing pad was $0.0787$ [m], with a maximum of $0.1192$ [m] and a minium of $0.0224$ [m]. Figure \ref{fig:solver} evaluates the real-time feasibility of our formulation. We measured the solver time per iteration on the onboard computer. The average computation time was $\approx4.9$ [ms], with all runs remaining well below the $10$ [ms] sampling period ($100$ [Hz]). This demonstrates that the optimization can be solved reliably in real time without exceeding hardware limits.

\section{CONCLUSION}
In this paper, we introduced a dynamic landing algorithm for MRUAVs that combines finite-horizon predictions with a fixed-time minimum-jerk trajectory planner to achieve precise synchronization with a moving landing pad and smooth, consistent descent. The framework effectively handles varying levels of measurement noise, which is essential for ensuring accuracy in such a demanding maneuver. Hardware experiments confirmed the robustness of the approach across multiple successful trials. Future work will focus on validating the framework in outdoor environments with wind disturbances and extending the MJT planner to incorporate energy constraints for more efficient landings.


\bibliographystyle{IEEEtran}
\bibliography{reference}

@IEEEtranBSTCTL{IEEEexample:BSTcontrol,
    CTLuse_forced_etal       = {yes},
    CTLmax_names_forced_etal = {2},
    CTLnames_show_etal       = {1}
}

@article{zhu2023nonlinear,
  title={A nonlinear model predictive control based control method to quadrotor landing on moving platform},
  author={Zhu, Bingtao and Zhang, BingJun and Ge, Quanbo},
  journal={Cognitive Computation and Systems},
  volume={5},
  number={2},
  pages={118--131},
  year={2023},
  publisher={Wiley Online Library}
}

@inproceedings{li2023nonlinear,
  title={Nonlinear mpc for quadrotors in close-proximity flight with neural network downwash prediction},
  author={Li, Jinjie and Han, Liang and Yu, Haoyang and Lin, Yuheng and Li, Qingdong and Ren, Zhang},
  booktitle={2023 62nd IEEE Conference on Decision and Control (CDC)},
  pages={2122--2128},
  year={2023},
  organization={IEEE}
}

@inproceedings{zhao2009adaptive,
  title={An adaptive UKF with noise statistic estimator},
  author={Zhao, Lin and Wang, Xiaoxu},
  booktitle={2009 4th IEEE Conference on Industrial Electronics and Applications},
  pages={614--618},
  year={2009},
  organization={IEEE}
}

@inproceedings{xing2011adaptive,
  title={An adaptive UKF algorithm and its application for satellite attitude determination},
  author={Xing, Ming and Hongzhuan, Qiu},
  booktitle={2011 International Conference on Consumer Electronics, Communications and Networks (CECNet)},
  pages={78--81},
  year={2011},
  organization={IEEE}
}

@article{borowczyk2017autonomous,
  title={Autonomous landing of a quadcopter on a high-speed ground vehicle},
  author={Borowczyk, Alexandre and Nguyen, Duc-Tien and Nguyen, Andr{\'e} Phu-Van and Nguyen, Dang Quang and Saussi{\'e}, David and Le Ny, Jerome},
  journal={Journal of Guidance, Control, and Dynamics},
  volume={40},
  number={9},
  pages={2378--2385},
  year={2017},
  publisher={American Institute of Aeronautics and Astronautics}
}

@article{guo2022autonomous,
  title={Autonomous landing of a quadrotor on a moving platform via model predictive control},
  author={Guo, Kaiyang and Tang, Pan and Wang, Hui and Lin, Defu and Cui, Xiaoxi},
  journal={Aerospace},
  volume={9},
  number={1},
  pages={34},
  year={2022},
  publisher={MDPI}
}

@article{ghommam2017autonomous,
  title={Autonomous landing of a quadrotor on a moving platform},
  author={Ghommam, Jawhar and Saad, Maarouf},
  journal={IEEE Transactions on Aerospace and Electronic Systems},
  volume={53},
  number={3},
  pages={1504--1519},
  year={2017},
  publisher={IEEE}
}

@article{feng2018autonomous,
  title={Autonomous landing of a UAV on a moving platform using model predictive control},
  author={Feng, Yi and Zhang, Cong and Baek, Stanley and Rawashdeh, Samir and Mohammadi, Alireza},
  journal={Drones},
  volume={2},
  number={4},
  pages={34},
  year={2018},
  publisher={MDPI}
}

@article{qi2019autonomous,
  title={Autonomous landing solution of low-cost quadrotor on a moving platform},
  author={Qi, Yuhua and Jiang, Jiaqi and Wu, Jin and Wang, Jianan and Wang, Chunyan and Shan, Jiayuan},
  journal={Robotics and Autonomous Systems},
  volume={119},
  pages={64--76},
  year={2019},
  publisher={Elsevier}
}

@article{gautam2022autonomous,
  title={Autonomous quadcopter landing on a moving target},
  author={Gautam, Alvika and Singh, Mandeep and Sujit, Pedda Baliyarasimhuni and Saripalli, Srikanth},
  journal={Sensors},
  volume={22},
  number={3},
  pages={1116},
  year={2022},
  publisher={Multidisciplinary Digital Publishing Institute}
}

@inproceedings{paris2020dynamic,
  title={Dynamic landing of an autonomous quadrotor on a moving platform in turbulent wind conditions},
  author={Paris, Aleix and Lopez, Brett T and How, Jonathan P},
  booktitle={2020 IEEE International Conference on Robotics and Automation (ICRA)},
  pages={9577--9583},
  year={2020},
  organization={IEEE}
}

@article{serra2016landing,
  title={Landing of a quadrotor on a moving target using dynamic image-based visual servo control},
  author={Serra, Pedro and Cunha, Rita and Hamel, Tarek and Cabecinhas, David and Silvestre, Carlos},
  journal={IEEE Transactions on Robotics},
  volume={32},
  number={6},
  pages={1524--1535},
  year={2016},
  publisher={IEEE}
}

@article{prochazka2024model,
  title={Model predictive control-based trajectory generation for agile landing of unmanned aerial vehicle on a moving boat},
  author={Proch{\'a}zka, Ond{\v{r}}ej and Nov{\'a}k, Filip and B{\'a}{\v{c}}a, Tom{\'a}{\v{s}} and Gupta, Parakh M and P{\v{e}}ni{\v{c}}ka, Robert and Saska, Martin},
  journal={Ocean Engineering},
  volume={313},
  pages={119164},
  year={2024},
  publisher={Elsevier}
}

@article{xuan2022quadcopter,
  title={Quadcopter precision landing on moving targets via disturbance observer-based controller and autonomous landing planner},
  author={Xuan-Mung, Nguyen and Nguyen, Ngoc Phi and Nguyen, Tan and Pham, Dinh Ba and Vu, Mai The and Thanh, Ha Le Nhu Ngoc and Hong, Sung Kyung},
  journal={IEEE Access},
  volume={10},
  pages={83580--83590},
  year={2022},
  publisher={IEEE}
}

@article{wang2022quadrotor,
  title={Quadrotor autonomous landing on moving platform},
  author={Wang, Pengyu and Wang, Chaoqun and Wang, Jiankun and Meng, Max Q-H},
  journal={Procedia Computer Science},
  volume={209},
  pages={40--49},
  year={2022},
  publisher={Elsevier}
}

@inproceedings{falanga2017vision,
  title={Vision-based autonomous quadrotor landing on a moving platform},
  author={Falanga, Davide and Zanchettin, Alessio and Simovic, Alessandro and Delmerico, Jeffrey and Scaramuzza, Davide},
  booktitle={2017 IEEE International Symposium on Safety, Security and Rescue Robotics (SSRR)},
  pages={200--207},
  year={2017},
  organization={IEEE}
}

@article{wang2021efficient,
  title={Efficient nonlinear model predictive control for quadrotor trajectory tracking: Algorithms and experiment},
  author={Wang, Dong and Pan, Quan and Shi, Yang and Hu, Jinwen and Zhao, Chunhui},
  journal={IEEE Transactions on Cybernetics},
  volume={51},
  number={10},
  pages={5057--5068},
  year={2021},
  publisher={IEEE}
}

@article{nan2022nonlinear,
  title={Nonlinear MPC for quadrotor fault-tolerant control},
  author={Nan, Fang and Sun, Sihao and Foehn, Philipp and Scaramuzza, Davide},
  journal={IEEE Robotics and Automation Letters},
  volume={7},
  number={2},
  pages={5047--5054},
  year={2022},
  publisher={IEEE}
}

@article{romero2022model,
  title={Model predictive contouring control for time-optimal quadrotor flight},
  author={Romero, Angel and Sun, Sihao and Foehn, Philipp and Scaramuzza, Davide},
  journal={IEEE Transactions on Robotics},
  volume={38},
  number={6},
  pages={3340--3356},
  year={2022},
  publisher={IEEE}
}

@INPROCEEDINGS{izadi2024multi,

  author={Izadi, Mohammadreza and Shayan, Zeinab and Faieghi, Reza},

  booktitle={2024 IEEE 20th International Conference on Automation Science and Engineering (CASE)}, 

  title={Multi-Model Predictive Attitude Control of Quadrotors}, 

  year={2024},

  volume={},

  number={},

  pages={3830-3835},

  doi={10.1109/CASE59546.2024.10711426}}

@misc{sun2022comparative,
      title={A Comparative Study of Nonlinear MPC and Differential-Flatness-Based Control for Quadrotor Agile Flight}, 
      author={Sihao Sun and Angel Romero and Philipp Foehn and Elia Kaufmann and Davide Scaramuzza},
      year={2022},
      eprint={2109.01365},
      archivePrefix={arXiv},
      primaryClass={cs.RO}
}

@article{shayan2024nonlinear,
  title={Nonlinear model predictive control of tiltrotor quadrotors using feasible control allocation},
  author={Shayan, Zeinab and Cristobal, Jann and Izadi, Mohammadreza and Yazdanshenas, Amin and Naderi, Mehdi and Faieghi, Reza},
  journal={Journal of Intelligent \& Robotic Systems},
  volume={111},
  number={2},
  pages={54},
  year={2025},
  publisher={Springer}
}

@article{tajeddin2019ecological,
  title={Ecological adaptive cruise control with optimal lane selection in connected vehicle environments},
  author={Tajeddin, Sadegh and Ekhtiari, Sanaz and Faieghi, Mohammadreza and Azad, Nasser L},
  journal={IEEE Transactions on Intelligent Transportation Systems},
  volume={21},
  number={11},
  pages={4538--4549},
  year={2019},
  publisher={IEEE}
}

@article{piazzi2000global,
  title={Global minimum-jerk trajectory planning of robot manipulators},
  author={Piazzi, Aurelio and Visioli, Antonio},
  journal={IEEE transactions on industrial electronics},
  volume={47},
  number={1},
  pages={140--149},
  year={2000},
  publisher={IEEE}
}

@article{verschueren2022acados,
  title={acados—a modular open-source framework for fast embedded optimal control},
  author={Verschueren, Robin and Frison, Gianluca and Kouzoupis, Dimitris and Frey, Jonathan and Duijkeren, Niels van and Zanelli, Andrea and Novoselnik, Branimir and Albin, Thivaharan and Quirynen, Rien and Diehl, Moritz},
  journal={Mathematical Programming Computation},
  volume={14},
  number={1},
  pages={147--183},
  year={2022},
  publisher={Springer}
}

@incollection{aswal2022joint,
  title={Joint and dual estimation of states and parameters with extended and unscented Kalman filters},
  author={Aswal, Neha and Bhattacharya, Baidurya and Sen, Subhamoy},
  booktitle={Recent Developments in Structural Health Monitoring and Assessment--Opportunities and Challenges: Bridges, Buildings and Other Infrastructures},
  pages={223--252},
  year={2022},
  publisher={World Scientific}
}

@article{shayan2025exponential,
  title={Exponential control barrier function and model predictive control for jerk-level reactive motion planning of quadrotors},
  author={Shayan, Zeinab and Izadi, Mohammadreza and Scognamiglio, Vincenzo and D’Angelo, Simone and Singoji, Shashank and Lippiello, Vincenzo and Faieghi, Reza},
  journal={Control Engineering Practice},
  volume={164},
  pages={106489},
  year={2025},
  publisher={Elsevier}
}

\balance
\end{document}